# Scalable Block-Diagonal Locality-Constrained Projective Dictionary Learning

**Zhao Zhang**[1], **Weiming Jiang**[2], **Zheng Zhang**[3], **Sheng Li**[4], **Guangcan Liu**[5] and **Jie Qin**[6]

[1] School of Computer Science & School of Artificial Intelligence, Hefei University of Technology, China
[2] School of Computer Science and Technology, Soochow University, China
[3] School of Information Technology and Electrical Engineering, University of Queensland, Australia
[4] Department of Computer Science, University of Georgia, USA
[5] School of Information and Control, Nanjing University of Information Science and Technology, China
[6] Computer Vision Laboratory, ETH Zurich, Switzerland

{cszzhang, cswmjiang}@gmail.com, darren.zhang@uq.edu.au, sheng.li@uga.edu, gcliu@nuist.edu.cn, jqin@vision.ee.ethz.ch

## Abstract

We propose a novel structured discriminative block-diagonal dictionary learning method, referred to as scalable Locality-Constrained Projective Dictionary Learning (LC-PDL), for efficient representation and classification. To improve the scalability by saving both training and testing time, our LC-PDL aims at learning a structured discriminative dictionary and a block-diagonal representation without using costly $l_0/l_1$-norm. Besides, it avoids extra time-consuming sparse reconstruction process with the well-trained dictionary for new sample as many existing models. More importantly, LC-PDL avoids using the complementary data matrix to learn the sub-dictionary over each class. To enhance the performance, we incorporate a locality constraint of atoms into the DL procedures to keep local information and obtain the codes of samples over each class separately. A block-diagonal discriminative approximation term is also derived to learn a discriminative projection to bridge data with their codes by extracting the special block-diagonal features from data, which can ensure the approximate coefficients to associate with its label information clearly. Then, a robust multiclass classifier is trained over extracted block-diagonal codes for accurate label predictions. Experimental results verify the effectiveness of our algorithm.

## 1 Introduction

Sparse Representation (SR) has delivered impressive results in applications of pattern recognition and data mining [Tropp and Gillert, 2010]; Jiang *et al.*, 2016; Zhang *et al.*, 2016; [Elad and Aharon, 2006]. SR by dictionary learning has been widely-used for reconstructing data by a linear combination of a few items in a dictionary [Jiang *et al.*, 2013]. That is, the dictionary plays a crucial role in the process of SR. But how to obtain a desired dictionary from data is still challenging.

Due to the fact that most high-dimensional real data can be sparsely represented by the representative samples of a class in a lower-dimensional manifold [Wright *et al.*, 2009], the sparse $l_0/l_1$-norm constraint on coefficients has been widely used in existing Dictionary Learning (DL) settings [Mairal *et al.,* 2010; Jiang *et al*., 2016; Zhang and Li, 2010; Li *et al.*, 2019]. Existing algorithms can be divided into unsupervised and supervised ones [Jiang *et al*., 2013]. Unsupervised DL models focus on minimizing the sparse reconstruction error to obtain a dictionary, e.g., [Wright *et al.*, 2009; Aharon *et al.*, 2006; Wang *et al.*, 2010], among which *K-Singular Value Decomposition* (KSVD) is one most representative method to generalize *K*-means clustering and obtain an over-complete dictionary from whole training set [Aharon *et al.*, 2006]. Note that KSVD only requires that the learned dictionary to well reconstruct the training data, thus it is not suitable for classification. Based on SR, [Wright *et al.*, 2009] proposed a *Sparse Representation-Based Classification* (SRC) method that uses entire training set as dictionary to learn the sparse codes by $l_1$-norm minimization. But, obtaining codes from a large dictionary is computationally expensive. It is also worth noting that some researches [Jiang *et al.* 2013; Gu *et al*.2014] have demonstrated that dictionary constructed via supervised learning tends to yield better discriminative performance.

One popular strategy for the supervised DL methods is to compute a shared dictionary for all classes. To enhance the classification performance, most existing models incorporate certain discriminative information to force the coefficients to be discriminating [Jiang *et al.,* 2013; Yang *et al.*, 2014]. For example, a classification error was incorporated into KSVD so that classification and representation performance can be jointly improved [Zhang and Li, 2010]. In [Jiang *et al*., 2013], a binary sparse code matrix over class labels was introduced to encourage intra-class samples to have similar sparse codes. The other supervised type is to learn a structured dictionary to promote discrimination between classes [Gu *et al*., 2014; Yang *et al*., 2014; Ramirez *et al*., 2010]. [Gu *et al*., 2014] extended the conventional synthesis dictionary learning to joint synthesis and analysis dictionary pair learning. [Yang et al., 2014] used a Fisher discrimination criterion with $l_1$-norm sparsity for the sparse coding to enhance the discriminating powers of learnt dictionary and coding coefficients.

It is worth noting that aforementioned methods still suffer from certain shortcomings. First, existing methods usually use the costly $l_0$ or $l_1$-norm sparsity constraint to obtain sparse codes for representation. But the use of $l_0/l_1$-norm constraint brings huge computation burden and thus makes the training phase inefficient. Second, for classification existing methods usually have two separable steps: coding and classification. Concretely speaking, given a new test sample, they have to obtain its coding coefficient using trained dictionary firstly under a sparsity constraint, and then classify it based on the pre-obtained codes, which is not straightforward and costly. It is worth noting that *Projective Dictionary Pair Learning* (DPL) [Gu et al., 2014] computes an analysis dictionary to obtain group sparse coefficients without $l_0/l_1$-norm constraint, and it can reduce the time complexity in training phase in the case that the training set is small. But it uses complementary data matrix of the data $X_l$ of each class $l$ in training set $X$ to learn the sub-dictionary $D_l$ of each class $l$, which means that lots of training time are needed when the size of training set is large. In addition, DPL did not consider the classification issue to train a classifier jointly, i.e., it is indeed a two-stage model, hence its performance may be degraded in reality.

In this paper, we propose a new scalable DL framework to overcome the drawbacks mentioned above. The major contributions of this paper are described as follows:

(1) A novel scalable block-diagonal Locality-Constrained Projective Dictionary Learning (LC-PDL) approach is technically proposed. LC-PDL calculates a structured dictionary and coefficients of each class separately by optimizing each sub-dictionary to reconstruct the intra-class data to obtain the block-diagonal representations. To improve the scalability, we clearly avoid using costly $l_0/l_1$-norm and complementary data matrix when seeking the sub-dictionary $D_l$ of each class $l$, which can potentially save lots of training time in large-scale cases. Besides, we avoid using extra time-consuming sparse reconstruction process with well-trained dictionary for each new sample to save testing time. Since real data usually contain noise and outliers, to preserve the locality, we include a Laplacian regularization over learned dictionaries to encourage "close" points to have similar codes [Li et al., 2015].

(2) To force the approximate codes to associate with label information and enforce the approximate codes to satisfy the block-diagonal structures as much as possible, we include a block-diagonal discriminative approximation term. LC-PDL also computes a projection to extract block-diagonal features to approximate the coefficients of samples.

(3) To minimize the classification error jointly, LC-PDL trains a robust $l_{2,1}$-norm based classifier based on approximate block-diagonal codes for accurate label predictions.

## 2 DPL Revisited

Suppose that $X=[X_1,\cdots,X_c]\in\mathbb{R}^{n\times N}$ is the training set, where $c$ is the total number of classes, $n$ is the original dimension and $N$ is the number of training samples. $X_i=[x_{i1},\cdots x_{iN_i}]\in\mathbb{R}^{n\times N_i}$ is the training subset of class $i$, where $N_i$ is the number of the training samples in class $i$. DPL aims at learning a synthesis class-specific dictionary $D=[D_1,\cdots D_c]\in\mathbb{R}^{n\times K}$ and an analysis dictionary $P=[P_1;\cdots P_c]\in\mathbb{R}^{K\times n}$ jointly by solving

$$\langle P,D\rangle=\arg\min_{P,D}\sum_{i=1}^{c}\left\|X_i-D_iP_iX_i\right\|_F^2+\lambda\left\|P_i\overline{X_i}\right\|_F^2, \ s.t.\left\|d_j\right\|_2^2\leq1, \quad (1)$$

where $\overline{X_i}$ is the complementary data matrix of $X_l$ in $X$, $d_j$ is the $j$-th atom of dictionary, $K$ is the number of atoms, the constraint $\left\|d_j\right\|_2^2\leq1$ over $d_j$ is to avoid the trivial solution of $P_i=0$ and can make the method stable. By setting $P_iX_q\approx0$, $\forall q\neq i$, the approximate codes $PX$ are nearly block-diagonal. It is worth noting that when the size of training set is large, the above computation burden will be heavy consequently.

Given a new test sample $x_{new}$, the recent DPL algorithm predicts its class label by

$$identity(x_{new})=\arg\min_l\left\|x_{new}-D_iP_ix_{new}\right\|_2. \quad (2)$$

That is, DPL classifies each new data $x_{new}$ by minimizing the residual between the testing sample and its reconstructed approximation by applying the synthesis sub-dictionary and the analysis sub-dictionary within each class, while it cannot obtain an explicit classifier for data classification.

## 3 Scalable Block-Diagonal Locality-Constrained Projective Dictionary Learning (LC-PDL)

### 3.1 The objective function

RDDL mainly improves the scalability of DL, and forces the approximate codes to deliver block-diagonal structures. To formulate the problem, we consider three sub-problems:

**Scalable structured block-diagonal projective DL.**
LC-PDL aims to learn the coding coefficients and dictionary of each class separately, so we first incorporate a structured synthesis dictionary $D=[D_1,\cdots D_c]\in\mathbb{R}^{n\times K}$. To gain a projection $P$ for extracting the codes from data, we define the scalable projective dictionary learning process as follows:

$$\min_{P,D}\sum_{i=1}^{c}\left\{\left\|X_i-D_iA_i\right\|_F^2+\tau\left\|PX_i-Q_iA_i\right\|_F^2\right\}$$
$$s.t.\left\|d_j\right\|_2^2\leq1, A_{j,l}\geq0, j\in\{1,\cdots,k\}, l=\{1,\cdots,N_i\} \quad (3)$$

where $X_i$ is the set of training samples of class $i$, $A_{j,l}\geq0$ is a nonnegative constraint, $N_i$ is the number of training samples in the class $i$ and $k$ is the number of the atoms in each class. $Q=[Q_1,\cdots Q_c]\in\mathbb{R}^{K\times K}$ is a discriminative transformation matrix and $Q_i\in\mathbb{R}^{K\times k}$ corresponds to the sub-dictionary in the class $i$. Denote the $m$-th row of $Q_i$ by $Q_i^m$, if $Q_i^m$ and the dictionary item $d_j$ share the same label, the elements in $Q_i^j$ are all ones; otherwise, $Q_i^j=0$. For example, assuming that $D=[D_1,\cdots D_6]$ and $X=[x_1,\cdots x_6]$, where $x_1$ and $x_2$ are from class 1, $x_3$ and $x_4$ are from class 2, $x_5$ and $x_6$ are from class 3, each sub-dictionary $D_i=[d_{2i-1},d_{2i}]$, then $Q$ can be defined as the euqation in (4).

Term $\left\|PX_i-Q_iA_i\right\|_F^2$ is discriminative approximation error, which enforces the approximate coding coefficients $PX_i$ to approximate the block-diagonal coefficients $Q_iA_i$. By forcing the approximate coding coefficients to associate with label information, one can enforce $PX$ to be block-diagonal.

$$Q = \begin{bmatrix} 1 & 1 & 0 & 0 & 0 & 0 \\ 1 & 1 & 0 & 0 & 0 & 0 \\ 0 & 0 & 1 & 1 & 0 & 0 \\ 0 & 0 & 1 & 1 & 0 & 0 \\ 0 & 0 & 0 & 0 & 1 & 1 \\ 0 & 0 & 0 & 0 & 1 & 1 \end{bmatrix}. \quad (4)$$

$\underbrace{\phantom{xx}}_{Q_1}\underbrace{\phantom{xx}}_{Q_2}\underbrace{\phantom{xx}}_{Q_3}$

**Locality preservation of atoms for codes.**
Encoding the pairwise similarities in DL process is important, as the locality constraint can ensure the learnt representation of samples to deliver similar structures. Following [Li *et al.*, 2015], we define an adjacent matrix $M$ over dictionary as

$$M_{v,j} = \begin{cases} \exp(-\|d_v - d_j\|_2 / \delta), & \text{if } d_j \in \mathbb{N}(d_v) \\ 0, & \text{otherwise} \end{cases}, \quad (5)$$

where $\delta$ is kernel width and $\mathbb{N}(d_v)$ is the $k$-nearest neighbor set of atom $d_v$. Then, a graph Laplacian $L$ can be defined as

$$L = M_d - M, \quad (6)$$

where $M_d$ is a diagonal matrix satisfying $(M_d)_{mm} = \sum_{l=1}^k M_{m,l}$. Since the profiles and atoms have one-to-one correspondence relationship, we can define the locality constraint as

$$\min_A \sum_{i=1}^c \left\{ \frac{1}{2} \sum_{j=1}^k \sum_{u=1}^k (a_j - a_u)^2 M_{j,u} \right\} = \sum_{i=1}^c Tr(A_i^T L_i A_i), \quad (7)$$

where $L_i$ is the graph Laplacian matrix of atoms in class $i$, $a_j$ and $a_u$ are the $j$-th row and $u$-th row of $A_i$, respectively. This locality constraint can ensure that the similar profiles encourage the corresponding atoms to be similar.

**Modeling a robust discriminartive classifier by applying block-diagonal representation.**
We introduce a variable label vector for each data $x_i$ as $h_i = [0,\cdots 1,\cdots 0] \in \mathbb{R}^c$, where the position of nonzero value indicates the label of $x_i$. We also assume that its label vector can be approximated by its embedded coefficient $Px_i$ using a linear classifier $W \in \mathbb{R}^{c \times K}$, $h_i \approx WPx_i$. Suppose that $H = [h_1,\cdots h_m] \in \mathbb{R}^{c \times m}$ are the class labels of training signals, we use the following $l_{2,1}$-norm regularized problem for robust classifier training:

$$\min_W \sum_{i=1}^N \|h_i - WPx_i\|_F^2 + \|W^T\|_{2,1} = \|H - WPX\|_F^2 + \|W^T\|_{2,1}, \quad (8)$$

where the $l_{2,1}$-norm regularization term $\|W^T\|_{2,1}$ can encourage the learned classifier $W$ to be sparse in rows [Zhang *et al.*, 2015] so that robust and discriminative soft labels can be learnt to improve the data representation and classification.

According to the above definitions in Eqs.(3), (7) and (8), we can obtain the final objective function of LC-PDL as

$$\min_{D,A,P,W} \sum_{i=1}^c \left\{ \|X_i - D_i A_i\|_F^2 + \tau \|PX_i - Q_i A_i\|_F^2 + \alpha Tr(A_i^T L_i A_i) \right\}$$
$$+ \beta \left( \|H - WPX\|_F^2 + \|W^T\|_{2,1} \right) \quad , \quad (9)$$
$$s.t. \|d_j\|_2^2 \leq 1, A_{j,l} \geq 0, j \in \{1,\cdots,k\}, l \in \{1,\cdots,N_i\}$$

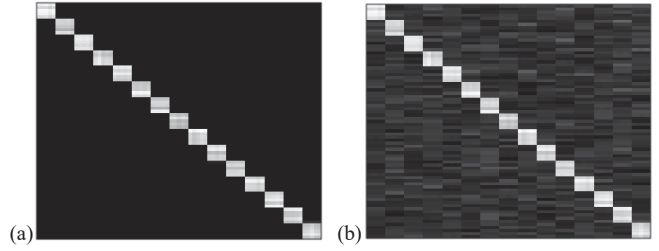

(a) (b)

Figure 1: Illustration of coding coefficients learned by using LC-PDL on the Yale face database. (a) The learned coding coefficients $A$; (b) The extracted coefficients $PX$ by direct emebdding.

where $\alpha \geq 0$ and $\beta \geq 0$ are tuning parameters. To illustrate that the learnt coding coefficients of training and test samples are block-diagonal, we prepare a simulation using Yale face database (at http://vision.csd.edu/content/yale-face-database). We firstly exhibit the coding coefficients $A$ using dictionary learning in Figure 1(a), and illustrate the approximate coding coefficients $PX$ by direct embedding with the projection $P$ in Figure 1 (b), from which we see that the learned coefficients matrix $A$ and approximate coefficients matrix $PX$ are all nearly block- diagonal, implying that the dictionary would be potentially overcomplete and the classification result over approximate coefficients $PX$ would be more accurate.

### 3.2 Optimization
We first initialize $D$, $P$ and $W$ to be random matrices with unit F-norm. The minimization can be alternated among the steps:

**Fix $D$ and $P$, optimize $A$.**
By removing the terms that are irrelevant to $A$, we can obtain the following sub-problem:

$$A^{(t+1)} = \arg\min_A \sum_{i=1}^c f(A_i), \; s.t. \; A^{(t+1)} \geq 0, \; where$$
$$f(A_i) = \|X_i - D_i^{(t)} A_i\|_F^2 + \tau \|P^{(t)} X_i - Q_i A_i\|_F^2 + \alpha Tr(A_i^T L_i A_i) \quad . \quad (10)$$

By setting $\partial f(A_i)/\partial A_i = 0$, we can compute $A^{(t+1)}$ as

$$A_i^{(t+1)} = \left( D_i^{(t)T} D_i^{(t)} + \tau Q_i^T Q_i + \alpha L_i / 2 + \alpha L_i / 2 \right)^{-1} \left( \tau Q_i^T P^{(t)} X_i + D_i^{(t)T} X_i \right),$$
$$A^{(t+1)} = \max(A^{(t+1)}, 0)$$
(11)

where $A^{(t+1)} = \max(A^{(t+1)}, 0)$ means that replacing the negative entries with 0 to ensure the nonnegative property of $A^{(t+1)}$.

**Fix $A$ and $W$, optimize $P$.**
By removing terms that are irrelevant to $P$, we can obtain the following sub-problem:

$$P^{(t+1)} = \arg\min_P g(P), \; where$$
$$g(P) = \beta \|H - WPX\|_F^2 + \sum_{i=1}^c \tau \|PX_i - Q_i A_i^{(t+1)}\|_F^2 \quad . \quad (12)$$

We consider to solve the following approximate problem:

$$P^{(t+1)} = \arg\min_P g(P), \; where$$
$$g(P) = \beta \|H - WPX\|_F^2 + \tau \|P[X_1,\cdots,X_c] - [Q_1 A_1^{(t+1)},\cdots Q_c A_c^{(t+1)}]\|_F^2 \quad .(13)$$
$$= \beta \|H - WPX\|_F^2 + \tau \|PX - M\|_F^2$$

By setting $\partial g(P)/\partial P = 0$, we can update projection $P^{(t+1)}$ as

$$P^{(t+1)} = \left(\tau I + \beta W^{(t)\mathrm{T}} W^{(t)}\right)^{-1} \left(\tau M^{(t+1)} X^{\mathrm{T}} + \beta W^{(t)\mathrm{T}} H X^{\mathrm{T}}\right)\left(X X^{\mathrm{T}}\right)^{-1}. \quad (14)$$

**Fix $P$, optimize $W$ and $\Lambda$.**
We update $W$ and $\Lambda$ alternately. According to property of the $l_{2,1}$-norm [Hou *et al.*, 2014; Zhang *et al.*, 2015], that is, $\|W^{\mathrm{T}}\|_{2,1} = 2tr(W\Lambda W^{\mathrm{T}})$, where $\Lambda$ is a diagonal matrix with the entries being $\Lambda_{mm} = 1/2\|w_m\|_2$ and $w_m$ is the *m*-th column of $W$. Then, we can update $W$ and $\Lambda$ by

$$\left(W^{(t+1)}, \Lambda^{(t+1)}\right) = \arg\min_{W,\Lambda} \wp(W,\Lambda), \text{ where } \wp(W,\Lambda) = \left\|H - WP^{(t+1)}X\right\|_F^2 + \left\|W^{\mathrm{T}}\right\|_{2,1} = \left\|H - WP^{(t+1)}X\right\|_F^2 + 2tr\left(W\Lambda W^{\mathrm{T}}\right), \quad (15)$$

when each $w_m \neq 0$. We first fix $\Lambda$ to update $W$. By setting $\partial \wp(W,\Lambda)/\partial W = 0$, we can easily update $W^{(t+1)}$ as

$$W^{(t+1)} = HX^{\mathrm{T}}P^{(t+1)\mathrm{T}}\left(P^{(t+1)}XX^{\mathrm{T}}P^{(t+1)\mathrm{T}} + 2\Lambda^{(t)}\right)^{-1}. \quad (16)$$

Then, we can use $W$ to update $\Lambda$. According to [Zhang *et al.*, 2015], we can update the diagonal elements of $\Lambda^{(t+1)}$ as

$$\Lambda_{mm}^{(t+1)} = 1/\left[2\|w_m^{(t+1)}\|_2\right], \quad (17)$$

where $w_m^{(t+1)}$ is the *m*-th column of $W^{(t+1)}$. $\Lambda$ is initialized to be an identity matrix similarly as [Hou *et al.*, 2014; Zhang *et al.*, 2015] that shows that this choice generally works well.

**Fix $A$, optimize $D$.**
The optimization process of updating $D$ can be formulated as

$$D^{(t+1)} = \arg\min_D \sum_{i=1}^c \left\|X_i - D_i A_i^{(t+1)}\right\|_F^2, \text{ s.t. } \|d_j\|_2^2 \leq 1, \quad (18)$$

where a variable $S$ is involved [Gu *et al.*, 2014]. The optimal solution of Eq.(18) can be obtained by ADMM algorithm:

$$\begin{cases} D^{(t+1)(t'+1)} = \arg\min_D \left\|X - D^{(t+1)}A^{(t+1)}\right\|_F^2 + \rho\left\|D^{(t+1)} - S^{(t')} - T^{(t')}\right\|_F^2; \\ S^{(t'+1)} = \arg\min_S \rho\left\|D^{(t+1)(t'+1)} - S - T^{(t')}\right\|_F^2, \text{ s.t. } \|s_j\|_2^2 \leq 1; \\ T^{(t'+1)} = T^{(t')} + D^{(t+1)(t'+1)} - S^{(t'+1)}, \text{update } \rho \text{ if appropriate.} \end{cases} \quad (19)$$

Then we can update the graph Laplacian matrix $L_i^{(t+1)}$ by (8) and (9). We summarize the LC-PDL in Algorithm 1.

### 3.3 Convergence Analysis and Complexity
The objective function of LC-PDL in Eq.(9) is a bi-convex problem for $\{(D,P,\Lambda),(W,A,L)\}$. By fixing $D$, $P$ and $\Lambda$, the function is convex for $W$, $L$ and $A$. Similarly, by fixing $W$, $L$ and $A$, the resulted function is also convex for $D$, $P$ and $\Lambda$, since $L$ completely depends on $D$. Thus, the proposed optimization of LC-PDL is actually an alternate convex search (ACS) method [Gorski *et al.*, 2007]. Note that the convergence of this kind of problem has been well studied in [Gu *et al.*, 2014]. As we can achieve the optimal solutions of updating variables $A$, $W$, $P$, $D$, $L$ and $\Lambda$, our objective function has a general lower bound and the algorithm can be ensured to converge to a stationary point. It is empirically found that LC-PDL converges rapidly. Figure 2 shows the convergence curve of LC-PDL on MIT CBCL database [Weyrauch *et al.*, 2004]. We can easily find that the objective function value of our LC-PDL method drops quickly and converges to a small value after 10 iterations in most cases.

---

**Algorithm 1** Scalable Locality-Constrained Projective DL
**Input**: Training data matrix X and class label matrix $H$
**Parameters**: Parameters $\tau$, $\alpha$, $\beta$, and dictionary size $K$
**Output**: $D$, $A$, $P$, $W$.
1: Initialize $D^{(0)}$ and $P^{(0)}$ as random matrices with unit F-norm, and initialize the diagonal matrix $\Lambda = I$; $t=0$;
2: **while** not converge **do**
3: $\quad t \leftarrow t+1$;
4: $\quad$ Update the coding coefficients matrix $A^{(t+1)}$ by (11);
5: $\quad$ Update the projection matrix $P^{(t+1)}$ by (14);
6: $\quad$ Update the linear multi-class classifier $W^{(t+1)}$ by (16);
7: $\quad$ Update the diagonal matrix $\Lambda^{(t+1)}$ by (17);
8: $\quad$ Update the dictionary $D^{(t+1)}$ by solving (18);
9: **end while**
10: **return** solution

---

In the training phase of LC-PDL, $A$, $W$, $P$, $D$, $L$ and $\Lambda$ are updated alternately. Note that we update $A$, $D$, and $L$ class by class. The complexity of updating the variables $A_i$, $D_i$ and $L_i$ is $O(k^2n + k^3 + KnN_i)$, $O(T(cnN + c^3 + c^2n + n^2c))$ and $O(k^3)$ in each iteration, respectively, where $T$ is the iteration number in ADMM algorithm for updating $D$. The time complexity of updating $W$, $P$ and $\Lambda$ are $O(cnN + cKn + KnN + K^3)$, $O(K^2c + K^3 + KnN + KcN)$ and $O(n^2N)$ in each iteration, respectively.

### 3.4 Classification Approach
After the projection $P^*$ and classifier $W^*$ are obtained by our LC-PDL, we can perform classification to include new test data. Given a new test sample $x_{new}$, we first embed it onto the projection $P^*$ in the form of $P^*x_{new}$ to approximate its coding coefficient. Then, we can embed the approximate coding coefficient onto the linear classifier $W^*$ further in the form of $W^*P^*x_{new}$, i.e., the soft label vector $f_{new}$ can be obtained as

$$f_{new} = W^*P^*x_{new} \in \mathbb{R}^{c \times 1}, \text{ where } W^* \in \mathbb{R}^{c \times N}, P^* \in \mathbb{R}^{N \times n}, \quad (20)$$

where the position corresponding to biggest element in label vector $f_{new}$ determines the class assignment of $x_{new}$. That is, the hard label of $x_{new}$ is assigned as $\arg\max_{i \leq c}(f_{new})_i$, where $(f_{new})_i$ is the *i*-th entry of estimated soft label vector $f_{new}$.

## 4 Experimental Results and Analysis
We mainly evaluate our algorithm for data classification. The performance of LC-PDL is compared with several closely related SRC [Wright *et al.*, 2009], K-SVD [Aharon *et al.*, 2006], D-KSVD [Zhang and Li, 2010], LLC [Wang *et al.*, 2010], LC-KSVD [Jiang *et al.*, 2013], FDDL [Yang *et al.*, 2014], DPL [Gu *et al*., 2014] and LCLE-DL [Li *et al*., 2015]. We perform all the simulations on a PC with Intel (R) Core (TM) i3-4130 CPU @ 3.4 GHz 8G.

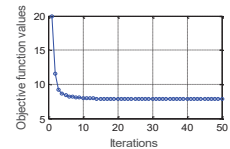

Figure 2: The convergence of our LC-PDL on MIT CBCL face database.

| Methods | MIT CBCL | AR | Caltech 101 | Caltech 256 |
|---|---|---|---|---|
| SRC | 93.1% | 66.5% | 48.3% | 42.6% |
| KSVD | 93.2% | 86.5% | 49.0% | 43.7% |
| DKSVD | 93.2% | 88.8% | 49.2% | 48.4% |
| LLC | 91.6% | 69.5% | 46.3% | 47.6% |
| LC-KSVD1 | 92.7% | 88.7% | 49.1% | 49.1% |
| LC-KSVD2 | 93.9% | 92.5% | 50.0% | 49.3% |
| LCLE-DL | 93.3% | 93.7% | 50.3% | 56.4% |
| FDDL | 96.0% | 95.6% | 49.6% | 52.7% |
| DPL | 94.7% | 95.8% | 48.4% | 73.5% |
| LC-PDL | **98.1**% | **97.3**% | **52.3**% | **78.0**% |

Table 1: Classification accuracies (%) on four image databases.

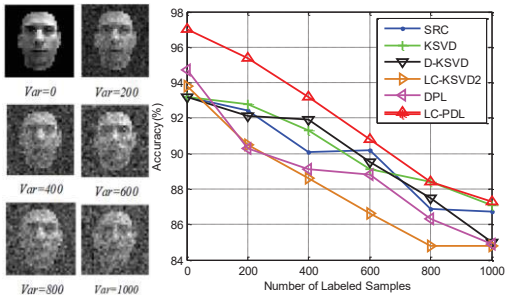

Figure 3: Classification performance with varying *Var* on MIT CBCL.

## 4.1 Image Classification

We provide the experimental results on image classification.

**MIT CBCL face database [Weyrauch *et al*., 2004]**
MIT CBCL database has 324 images per person (3,240 face images totally) rendered from 3D head models. Following the common evaluation procedures, the second face set is tested. All the face images are resized into 32×32 pixels for efficiency. Note that we first normalize each sample to have unit $l_2$-norm for simulations. We randomly select 4 images per person for training, while test on the rest. The number of dictionary atoms is simply set as the number of training data. Parameters $\tau$=0.01, $\alpha$=0.01 and $\beta$=0.1 is used in our LC-PDL.

**AR face database [Martinez and Benavente, 1998]**
The AR face database consists of over 4,000 color images of 126 people. Each person has 26 face images taken during two sessions. By following the standard evaluation procedures in [Jiang *et al*., 2013], the sample set that contains 2,600 images of 50 male subjects and 50 female subjects is selected for the simulations. Each face image of 165×120 pixels is projected onto a 540-dimensional vector with a generated matrix from a zero-mean normal distribution. By following [Jiang *et al*., 2013], we also randomly choose 20 images per person for training and the rest for testing. The dictionary contains 500 items, corresponding to an average of 5 items each category. $\tau$=0.01, $\alpha$=0.01, $\beta$=0.1 is set in our LC-PDL.

**Caltech101 database[Perona *et al*., 2004]**
The Caltech101 database contains 9144 images, consisting of 101 object classes and 1 background class. The number of images in each class varies from 31 to 800. By following the common evaluation procedure, the standard bag-of-words (BOW) plus spatial pyramid matching (SPM) [Lazebnik *et al*., 2006] approach is applied for feature extraction. Dense SIFT descriptors are extracted on the grid sizes of 1×1, 2×2 and 4×4 to obtain the SPM features. We then use the vector quantization based coding way to extract mid-level features and use the standard max pooling method to obtain the high-dimensional pooled features. At last, the dimension of pooled features is reduced to 3,000 by PCA. In this experiment, we randomly select 5 images from each category for training and use the rest for testing. The dictionary contains 204 items, corresponding to an average of 2 items each category. The parameters $\tau$=0.01, $\alpha$=0.1 and $\beta$=0.1 are set for LC-PDL.

**Caltech256 database [Griffin *et al*., 2007]**
We evaluate each method by using Caltech256 database that contains 30,607 images of 257 categories, such as bear, brain, tower, car, etc. The number of images in each class is at least 80. Each object image has been resized into 64×64 due to the consideration of computational efficiency. In this study, we perform dictionary learning using discriminant features, i.e., we use the *Linear Discriminant Analysis* (LDA) [Fukunaga, 1990] to reduce the dimension. For LDA, the eign-problem is solved by SVD, the regularization parameter is set to 0.1. Finally, the dimension of each image is reduced to 256. We randomly select 10 images per category for training and test on the rest. The dictionary size is set as its maximum for each algorithm. $\tau$=0.001, $\alpha$=0.1 and $\beta$=1e−4 are set in LC-PDL.

We repeat the experiments 10 times with different random spits of training and test images and average the results over different runs. All the classification results are illustrated in Table 1. We can find that our presented LC-PDL algorithm can deliver better accuracies than its competitors on all the used databases under the same setting.

## 4.2 Image Classification against Noisy Databases

We evaluate the robustness of our algorithm in this section. Technically, some numerical results have been provided to illustrate the robustness of our algorithm for handing the face images with noise. The MIT CBCL database is involved in this study, and the same setting as Subsection 4.1 is also used here. Random Gaussian noise under different variance *Var* is manually added to each image. We then employ the noisy databases for training and testing. We illustrate some noisy face images by setting *Var*=200, 400, 600, 800 and 1000, respectively. The experimental results are shown in Figure 3. From the classification results, we can easily find that the increasing of *Var* clearly decreases the performance of each method. Since the noise has been added to each image, some important information is lost (see Figure 3) and thus the recognition rates are decreased. We note that our algorithm delivers higher accuracies than others in most cases and the descending rate of our method is also comparable to the other baselines. The main reason is that a robust $l_{2,1}$-norm is used on the linear classifier, so that discriminative features can be obtained for more accurate label prediction results [Hou *et al*., 2014; Zhang *et al*., 2015].

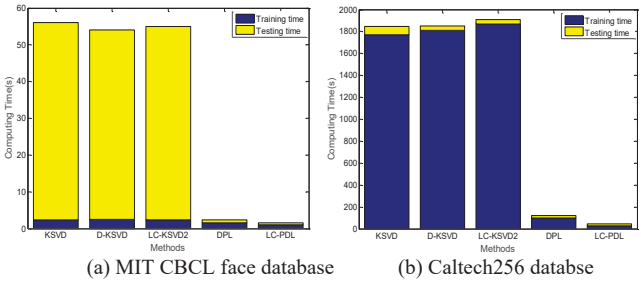

(a) MIT CBCL face database  (b) Caltech256 databse

Figure 4: The visualizations of needed time in training and testing phases.

### 4.3 Comparison of Computational Time

We use the MIT CBCL and Caltech256 databases to evaluate the runung time performance of each method in this study, and use the same setting as Subsection 4.1. That is, we use 40 and 2,570 images as the training set and use 3,200 and 28,037 images as the test set for MIT and Caltech256, respectively. The computational time (including the training and testing time) of each method is shown in Figure 4. We can find that when the size of training set is small, all evaluated methods spend less time in training on MIT. But for a large testing set, the testing time of KSVD, DKVD and LC-KSVD2 is far more than those of DPL and our LC-PDL, which is due to the fact that KSVD, DKSVD and LC-KSVD have involved extra time-consuming sparse reconstruction process for each test data to obtain the sparse codes for classification. From Figure 4b, we can observe that when the training set contains a large number of images, both the needed training time and testing time of KSVD, DKSVD, LC-KSVD2 and DPL are more than our LC-PDL, which is due to the fact that KSVD, DKSVD and LC-KSVD employ the time-consuming $l_0$ or $l_1$-norm sparsity constraint to obtain the codes, and DPL uses a large complementarty data matrix for the optimization, while our proposed LC-PDL uses a new discriminative approximation error term to avoid the above problems and it can force the approximate coding coefficients to be block-diagonal.

### 4.4 Parameter analysis

We analyze the parameter sensitivity of our algorithm in this study. Since parameter selection still remains an open issue, we also use a heuristic approach to select the most important parameters ($\tau$, $\alpha$, $\beta$) for our model. Specifically, we fix one of parameters and explore the effects of other two on the performance employing grid search. In this study, the AR face dataset is utilized. 20 face images of each person are randomly chosen as input signals. The rest is used for testing. The trained dictionary contains 500 items. For each pair of parameters, we average the results based on 10 random splits of training and testing images with varied parameters from $\{10^{-6}, 10^{-4}, \cdots 10^{6}\}$. We show the parameter selection results in Figure 5. LC-PDL works well in a wide range of parameters $\alpha$ and $\beta$, which means our LC-PDL model is insensitive to the parameters $\alpha$ and $\beta$ by delivering stable performances. It is also noted that that a larger $\tau$ than $10^{-2}$ tend to decrease the recognition result, i.e., a small $\tau$ can be used.

## 5 Concluding remarks

We proposed a new structured and scalable discriminative dictionary learning framework that integrates the dictionary learning, block-diagonal representation and classification to a unified model. To make the learnt dictionary discriminative for representation, we proposed an efficient way to consider local and label information of dictionary atoms to ensure the block-diagonal structure of coeffcients without using costly $l_0/l_1$-norm sparsity. We also incorporate a new discriminative approximation term to compute a projection to approximate the codes by extracting block-diagonal features of data, and train a classifier over approximate codes. This classification process is efficient since it avoids the extra time-consuming sparse reconstruction process with well-trained dictionary for each new data, suffered in many existing methods. LC-PDL also reduces the computation cost in the training phase by avoiding the invovlvement of the complementary data matrix when learning the sub-dictionary over each class.

We have examined the effectiveness of our algorithm on several widely-used image databases. The investigated cases demonstrate superior performances of our proposed method in classification and efficiency, by comparing with several related dictionary learning methods. In future, extending our method to the semi-supervised scenario needs investigation, since the number of labeled data is typically small in practice.

### Acknowledgments

The authors acknowledge support of the National Natural Science Foundation of China (Nos. 61672365, 61732008, 61725203, 61871444), and the Fundamental Research Funds for the Central Universities of China (JZ2019HGPA0102).

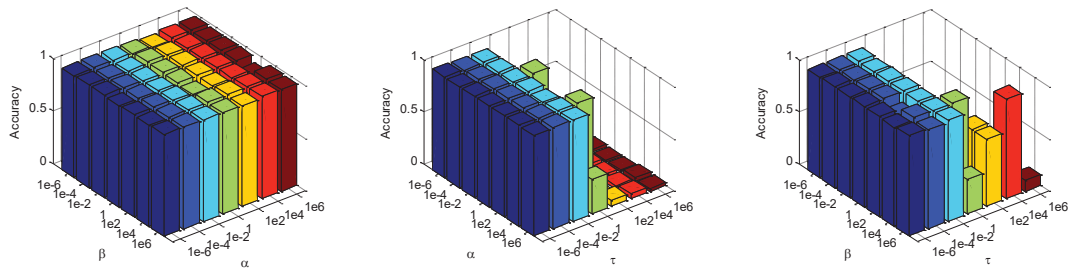

Figure 5: Parameter sensitivity analysis of our LC-PDL under different parameters on AR dataset. Left: the effects of tuning $\alpha$ and $\beta$ by fixing $\tau=0.01$; Middle: the effects of tuning $\alpha$ and $\tau$ by fixing $\beta=0.1$; Right: the effects of tuning $\beta$ and $\tau$ by fixing $\alpha=0.1$.